\documentclass[runningheads]{llncs}
\usepackage{graphicx}
\usepackage{times}  %Required
\usepackage{helvet}  %Required
\usepackage{courier}  %Required
\usepackage{url}  %Required
\usepackage{algorithm}
\usepackage{algorithmic}
\usepackage{multicol}
\usepackage{multirow}
\usepackage{subfigure}
\usepackage{bm}
\usepackage{appendix}
\usepackage{textcomp}

\begin{document}
\title{Auto-Encoder based Co-Training Multi-View Representation Learning \thanks{This work was supported by the National Key R\&D Program of China (No. 2016YFC0303703).}}

\author{Run-kun Lu\inst{1} \and
Jian-wei Liu\inst{1} \and
Yuan-fang Wang\inst{1} \and
Hao-jie Xie\inst{1} \and
Xin Zuo\inst{1}}

\authorrunning{Rk. Lu et al.}

\institute{Department of Automation, China University of Petroleum, Beijing 102249, China \\
\email{zsylrk@gmail.com}\\
\email{\{liujw, zuox\}@cup.edu.cn}}

\maketitle              % typeset the header of the contribution

\begin{abstract}
Multi-view learning is a learning problem that utilizes the various representations of an object to mine valuable knowledge and improve the performance of learning algorithm, and one of the significant directions of multi-view learning is sub-space learning. As we known, auto-encoder is a method of deep learning, which can learn the latent feature of raw data by reconstructing the input, and based on this, we propose a novel algorithm called Auto-encoder based Co-training Multi-View Learning (ACMVL), which utilizes both complementarity and consistency and finds a joint latent feature representation of multiple views. The algorithm has two stages, the first is to train auto-encoder of each view, and the second stage is to train a supervised network. Interestingly, the two stages share the weights partly and assist each other by co-training process. According to the experimental result, we can learn a well performed latent feature representation, and auto-encoder of each view has more powerful reconstruction ability than traditional auto-encoder.

\end{abstract}

\section{Introduction}
In real word applications, multi-view learning problems are widespread and they often exist in two ways. The first one is that multiple views exist naturally in data, such as we can easily obtain three views from web pages of Facebook, they include the content of the web page, the text of any web pages linking to this web page, and the link structure of all linked pages. The second one is that the raw data is not multi-view data and we need to construct multiple views for data, which include random approaches \cite{DBLP:conf/icdm/BickelS04,DBLP:conf/icml/BrefeldS04,DBLP:conf/ecml/BrefeldBS05} reshape or decompose approaches \cite{DBLP:journals/pr/WangCG11}, and the methods that perform feature set partitioning automatically \cite{DBLP:conf/icml/ChenWC11}. Once we get multiple views of raw data, we can utilize the advantages of multi-view learning to improve the performance of learning tasks like regression, classification and clustering, where multi-view learning methods can be classified into three categories: co-training, multiple kernel learning, and subspace learning \cite{DBLP:journals/corr/abs-1304-5634}. In this paper, we focus on subspace learning and propose a novel multi-view learning algorithm called Auto-encoder based Co-training Multi-View Learning (ACMVL) which utilizes both complementarity and consistency and finds a joint latent feature representation of multiple views. Note that co-training in our proposed algorithm's name is a training strategy instead of co-training multi-view learning method.

Multiple views of raw data have two wonderful properties, which are consistency and complementarity. Consistency represents the common information of multiple views, and complementarity represents the special information of each view. Only consistency and complementarity of multiple views can be utilized to improve the performance of learning tasks \cite{DBLP:conf/icml/ChaudhuriKLS09,DBLP:conf/icaisc/KursunA10,DBLP:conf/colt/BlumM98,Ou2018}, however, both consistency and complementarity are significant that it is a waste of information if we ignore one of them. In \cite{DBLP:journals/tnn/LiuJLZL15}, they find a joint latent representation which include both common and special features of multiple views, and followed this work, \cite{DBLP:conf/dasfaa/ZhangQLYS18} made some improvements. In their works, they compute the special feature of each view as well as the common feature of all views according to matrix factorization, and then concatenate them together in to a joint latent feature. However, this kind of method has two constraints:\\
(1) it is not reasonable to define all views share a common feature, maybe they share a common space and each view has its own instantiation in this space;\\
(2) the optimization algorithm is hard to adapt large scale data set, because the algorithm requires to feed all the training data instead of a batch at one time.
\begin{figure}[H]
\begin{center}
\includegraphics[width=0.8\textwidth]{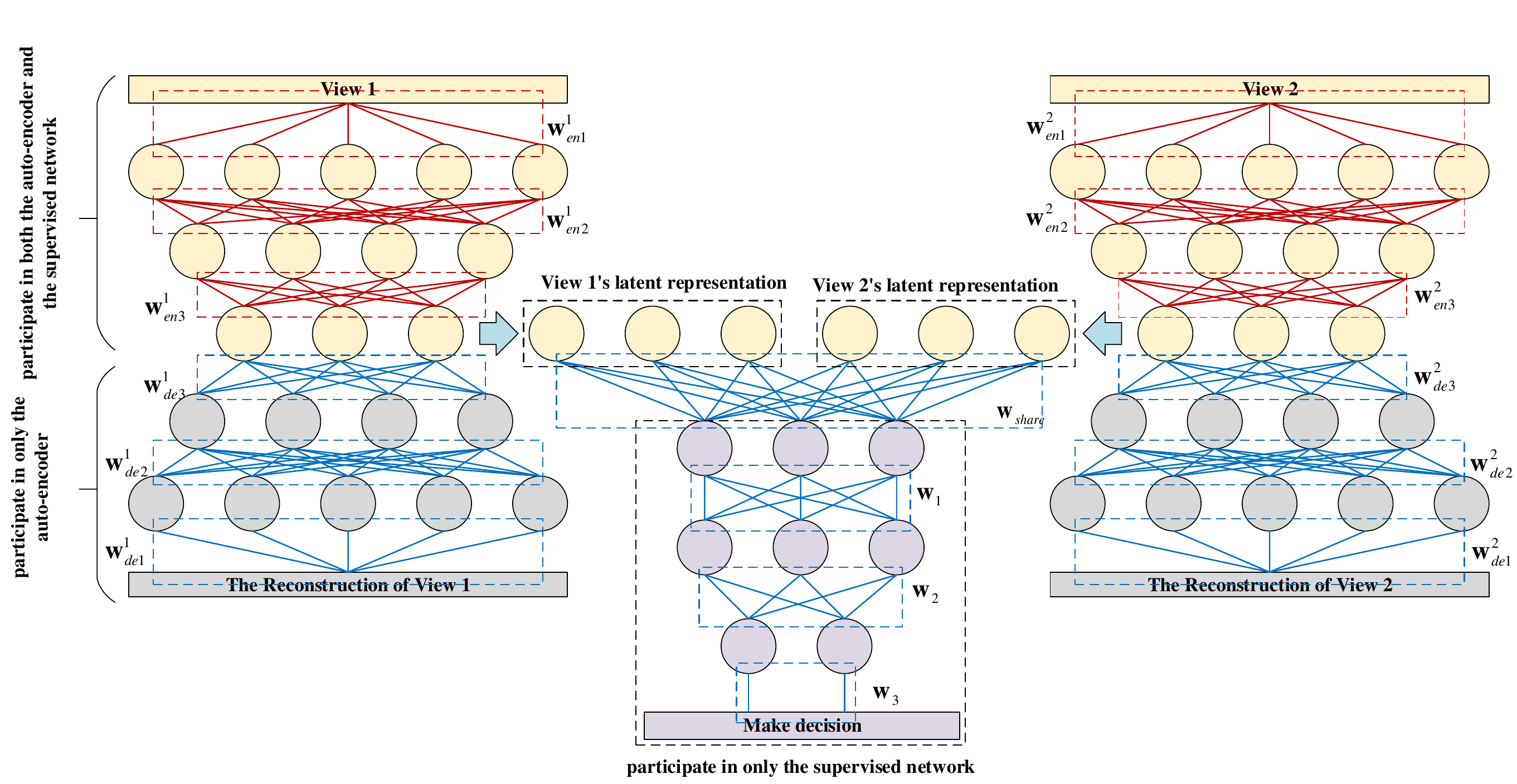}
\caption{Diagram of ACMVL: we illustrate a two-views' problem in this figure, therefore we have three networks. The left and right ones are View 1 and View 2's auto-encoder, and the middle one is the supervised network. Note that the yellow layers not only participate in the training process of auto-encoder but also supervised network, and our joint latent representation is the first purple layer of supervised network.} \label{splncs04}
\end{center}
\end{figure}
To solve such two problems, our proposed algorithm ACMVL builds a framework including multiple auto-encoders and a supervised network (a multi-layer perceptron which predicts labels of instances and minimize the cross-entropy loss between prediction and true label). Our proposed approach can easily solve the second question by running mini-batch gradient descent on large scale training sets. As for the first question, we first compute each view's special feature and then map each special feature into a same space by weight sharing and add them together, furthermore, with a nonlinear activation function we can get a joint latent representation. Besides, alternating co-training is another salient characteristic of ACMVL, which is a training strategy that lets our auto-encoders and supervised network partially share model parameters, and also let supervised network help auto-encoders meliorate their encoders' model parameters. And surprisingly, we find that by using this strategy, we can not only accelerate the convergence of the algorithm but also improve the learning performance of each auto-encoder significantly, which means each view will obtain a much better special feature that can help with the construction of joint latent feature. As a result, we will make a summarize of contributions we made as follows:\\
(1) We propose a novel multi-view learning algorithm ACMVL, which utilizes both consistency and complementarity to build a joint latent representation of multiple views, where multiple views' auto-encoders consider the consistency, and the weight sharing method considers the complementarity;\\
(2) Compared with the algorithms proposed by \cite{DBLP:journals/tnn/LiuJLZL15,DBLP:conf/dasfaa/ZhangQLYS18}, ACMVL is neural network-based algorithm which is easy to use mini-batch to adapt large scale data set.\\
(3) We propose an alternating co-training strategy which let our auto-encoders and supervised network partially shared model parameters and also let supervised network helps each auto-encoder to meliorate their encoder's weight. And this strategy can accelerate the training process and improve the learning performance of each auto-encoder significantly.

\section{Framework}
\subsection{Notations}
 In this paper, bold uppercase characters are used to denote matrices, bold lowercase characters are used to denote vectors, and other characters which are not bold are all used to denote scalars. Supposed that $({{\bf{X}}^v},{\bf{Y}})$ is the sample of view \emph{v}, where $v = 1, \cdots ,V$. Among of them, ${{\bf{X}}^v} \in {\Re^{{M^v} \times N}}$ is the set of input instances of view \emph{v}, ${\bf{Y}} \in {\Re^N}$ is the label, where \emph{N} is the number of instances, ${M^v}$ is the feature number of each instance of view \emph{v}. More specific, we have \emph{V} version of raw data, each version can be expressed as ${{\bf{X}}^v}$, and ${{\bf{X}}^v} = \left[ {{\bf{x}}_1^v, \cdots ,{\bf{x}}_N^v} \right]$, ${\bf{x}}_i^v \in {\Re^{{M^v}}}$. Note that all the views share the label ${\bf{Y}}$ because they are the various representation of raw input date, and ${\bf{Y}} = \left[ {{{\bf{y}}_1}, \cdots ,{{\bf{y}}_N}} \right]$, ${{\bf{y}}_i} \in \Re$.

\subsection{Core Concept of Framework}
As we known, auto-encoder is an algorithm that can compute the latent feature of raw data, and we can compute each view's latent feature by using auto-encoder. However, the \emph{V} views' latent features we obtained only consider the complementarity of different views, and we cannot guarantee that all of the auto-encoders can generate good latent features. Therefore, we aim to find a joint latent representation by combining the \emph{V} views' latent features we obtained according to some rules. In this paper, we build a simple multi-layer perceptron with its input of multiple views' latent features to supervise the process of the generation of joint latent representation. Furthermore, we adopt a novel training strategy to train multiple auto-encoders in each view as well as supervised network, we will give a description in detail in next subsection.

\subsection{Description of Framework}
Our proposed method ACMVL has a co-training process, which has two stages, one is the stage of learning latent feature that we need to train auto-encoders of multiple views, and the other is the stage of meliorating feature and learning joint latent feature that we need to train a supervised network. Next, we will explain the two scenarios separately.

\textbf{Latent feature learning}: as shown in Figure 1, we illustrate an example with two views. In Figure 1, there are two auto-encoders because we need to compute the latent feature of each view. Therefore, we need to train the model parameters to minimize the reconstruction error of each view, and we can formulate this problem as:
\begin{equation}
\min \frac{1}{V}{\sum\nolimits_{v = 1}^V {\left\| {{{\bf{X}}^v} - {{{\bf{\hat X}}}^v}} \right\|} _F}
\end{equation}
where ${{\bf{\hat X}}^v}$ is the reconstruction of view \emph{v}'s input. In each auto-encoder, the activation function is ReLu except for the last layer because the last layer of each auto-encoder is the reconstruction of raw data, and the optimizer algorithm we used is AdaDelta \cite{DBLP:journals/corr/abs-1212-5701}. After training two auto-encoders, we should save the model parameters $\theta _{en}^v = \left\{ {{\bf{w}}_{en1}^v,{\bf{w}}_{en2}^v,{\bf{w}}_{en3}^v} \right\}$, and $\theta _{de}^v = \left\{ {{\bf{w}}_{de1}^v,{\bf{w}}_{de2}^v,{\bf{w}}_{de3}^v} \right\}$, note that each view's auto-encoder has its own parameters.

\textbf{Meliorate feature and joint feature learning}: when finish the training process of auto-encoders, we take out the third layer of each auto-encoder as the input of the supervised network as shown in the middle of Figure 1. By mapping each latent feature representation of each view into a same subspace and add them together, we can easily find a joint latent representation by using a nonlinear mapping, which can be formulated as follows:
\begin{equation}
g\left( {\sum\nolimits_{v = 1}^V {{{\bf{w}}_{share}}{{\bf{h}}^v}} } \right)
\end{equation}
where $g\left(  \cdot  \right)$ is a nonlinear activation function, and we use ReLu in this paper. Note that we share the transform matrix ${{\bf{w}}_{share}}$ which helps us to find the consistent factors in various views. Furthermore, in this network, the objective function of supervised network is to minimize the cross-entropy loss between prediction and true label, which can be formulated as follows:
\begin{equation}
\min \frac{1}{N}\sum\nolimits_{i = 1}^N {\left( {{{\bf{y}}_i}\log \left( {{{{\bf{\hat y}}}_i}} \right) + \left( {1 - {{\bf{y}}_i}} \right)\log \left( {1 - {{{\bf{\hat y}}}_i}} \right)} \right)}
\end{equation}
where ${{\bf{\hat y}}_i}$ is the prediction of \emph{i}-th instance. As for the choice of activation function and optimizer algorithm, expect that last layer uses Softmax as activation function, others choose ReLu, and AdaDelta is selected to optimize the objective function. It is remarkable that not only we should train the parameters ${\theta _{sup}} = \left\{ {{{\bf{w}}_{share}},{{\bf{w}}_1},{{\bf{w}}_2},{{\bf{w}}_3}} \right\}$ as shown in Figure 1, but also need to update the value of $\theta _{en}^v = \left\{ {{\bf{w}}_{en1}^v,{\bf{w}}_{en2}^v,{\bf{w}}_{en3}^v} \right\}$, which is inherited from last stage, where $v = 1, \cdots ,V$. Same as last stage, we need to save parameter ${\theta _{sup}}$ as well as the updated parameter $\theta _{en}^v$.
\begin{figure}[ht]
\begin{center}
\includegraphics[width=0.8\textwidth]{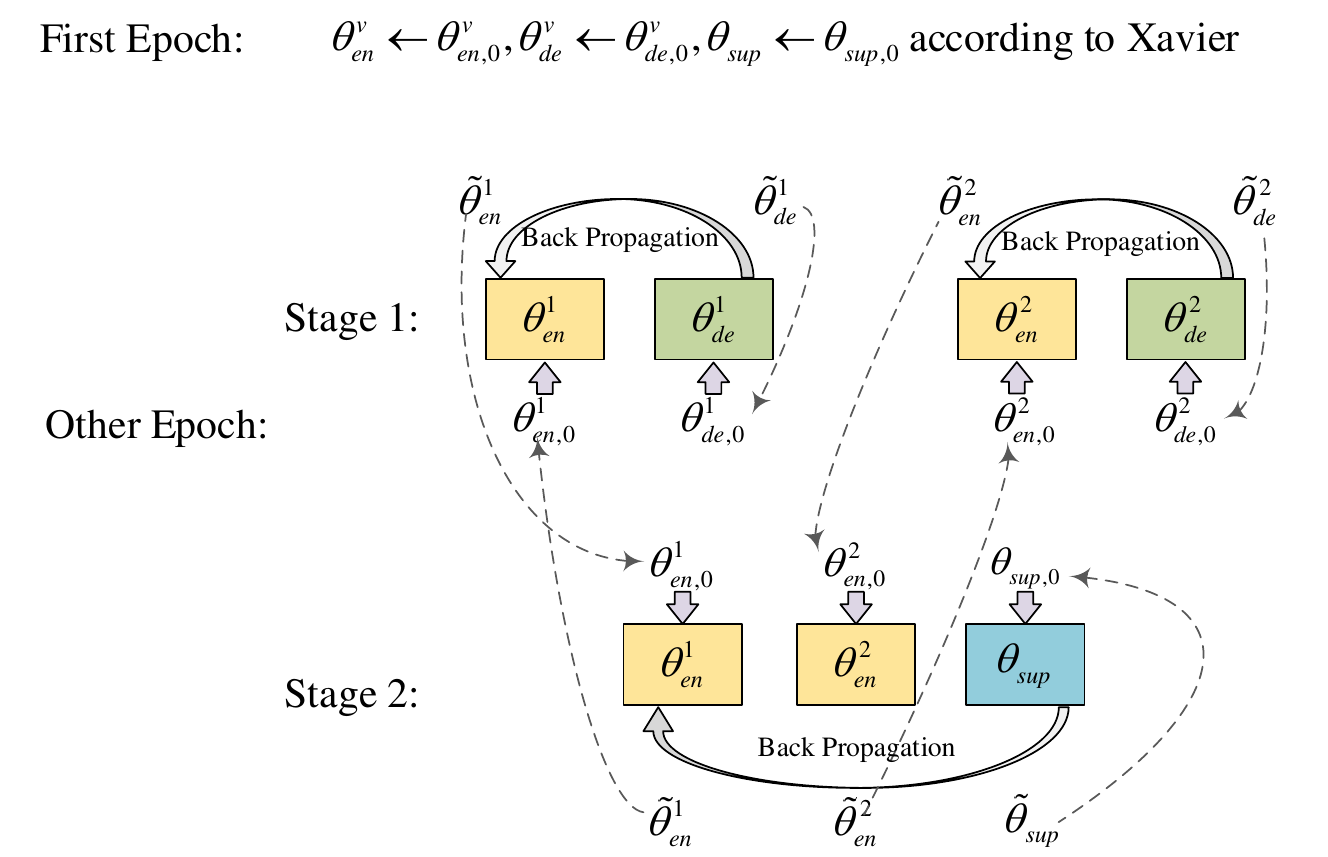}
\caption{Parameter Updating Process} \label{splncs04}
\end{center}
\end{figure}
The two stages we illustrate above is only just one epoch of training procedure, and we will design a co-training process. We define in one epoch, auto-encoder will be trained for ${R_1}$ rounds, and the supervised network will be trained for ${R_2}$ rounds. In first epoch, we need to initialize $\theta _{en}^v$, $\theta _{de}^v$, and ${\theta _{sup}}$ as $\theta _{en,0}^v$, $\theta _{de,0}^v$, and ${\theta _{sup,0}}$ according to the method of Xavier \cite{DBLP:journals/jmlr/GlorotB10}, and with these initial value we can conduct the first stage and obtain the best model parameters $\tilde \theta _{en}^v$ and $\tilde \theta _{de}^v$. Next, in stage 2, we use $\tilde \theta _{en}^v$ and ${\theta _{sup,0}}$ to initialize the supervised network and obtain the best value after ${R_2}$ rounds' training. Similarly, in other epochs, the training strategy only has minor difference that we do not initialize parameters according to Xavier. Specifically, in first stage, we initialize $\theta _{en}^v$ as $\tilde \theta _{en}^v$ from stage 2 of last epoch, and initialize $\theta _{de}^v$ as $\tilde \theta _{de}^v$ from stage 1 of last epoch; in second stage, we initialize $\theta _{en}^v$ as $\tilde \theta _{en}^v$ from stage 1 of this epoch, and initialize ${\theta _{sup}}$ as $\tilde \theta _{sup}^v$ from stage 2 of last epoch. To illustrate this process more clearly, we summarized the whole algorithm of ACMVL in Algorithm 1 and illustrate the parameter updating process of co-training in Figure 2.

\begin{algorithm}[tb]
\caption{ACMVL}
\label{alg:example}
\begin{algorithmic}
   \STATE Initialize $\theta _{en}^v$, $\theta _{de}^v$ and ${\theta _{sup}}$ as $\theta _{en,0}^v$, $\theta _{de,0}^v$, and ${\theta _{sup,0}}$ according to the method of Xavier;
   \STATE \textbf{For} each epoch \textbf{do}:
   \STATE \ \ \ \ \textbf{Stage 1}:
   \STATE \ \ \ \ \textbf{For} each view $v = 1, \cdots ,V$ \textbf{do}:
   \STATE \ \ \ \ \ \ \ \ \textbf{If} not first epoch:
   \STATE \ \ \ \ \ \ \ \ \ \ \ \ Initialize $\theta _{en}^v$ and $\theta _{de}^v$:
   \STATE \ \ \ \ \ \ \ \ \ \ \ \ $\theta _{en}^v:\theta _{en,0}^v \leftarrow \tilde \theta _{en}^v$, where $\tilde \theta _{en}^v$ comes from last\\
   \ \ \ \ \ \ \ \ epoch of stage 2;
   \STATE \ \ \ \ \ \ \ \ \ \ \ \ $\theta _{de}^v:\theta _{de,0}^v \leftarrow \tilde \theta _{de}^v$, where $\tilde \theta _{de}^v$ comes from last\\
   \ \ \ \ \ \ \ \ epoch of stage 1;
   \STATE \ \ \ \ \ \ \ \ \textbf{For} each ${R_1}$ \textbf{do}:
   \STATE \ \ \ \ \ \ \ \ \ \ \ \ Update $\theta _{en}^v$ and $\theta _{de}^v$ using AdaDelta and set the\\
   \ \ \ \ \ \ \ \ best one as $\tilde \theta _{en}^v$ and $\tilde \theta _{de}^v$ (select the weight of the\\
   \ \ \ \ \ \ \ \ round with the least reconstruction error);
   \STATE \ \ \ \ \ \ \ \ \textbf{End For}
   \STATE \ \ \ \ \textbf{End For}
   \STATE \ \ \ \ \textbf{Stage 2}:
   \STATE \ \ \ \ \textbf{If} not first epoch:
   \STATE \ \ \ \ \ \ \ \ Initialize $\theta _{en}^v$ and $\theta _{sup}^v$:
   \STATE \ \ \ \ \ \ \ \ $\theta _{en}^v:\theta _{en,0}^v \leftarrow \tilde \theta _{en}^v$, where $\tilde \theta _{en}^v$ comes from this\\
   \ \ \ \ epoch of stage 1;
   \STATE \ \ \ \ \ \ \ \ $\theta _{sup}^v:\theta _{sup,0}^v \leftarrow \tilde \theta _{sup}^v$, where $\tilde \theta _{sup}^v$ comes from last\\
   \ \ \ \ epoch of stage 2;
   \STATE \ \ \ \ \textbf{For} each ${R_2}$ \textbf{do}:
   \STATE \ \ \ \ \ \ \ \ Update $\theta _{en}^v$ and $\theta _{sup}^v$ using AdaDelta and set the\\
   \ \ \ \ best one as $\tilde \theta _{en}^v$ and $\tilde \theta _{sup}^v$ (select the weight of the\\
   \ \ \ \ round with the least reconstruction error);
   \STATE \ \ \ \ \textbf{End For}
   \STATE \textbf{End For}

\end{algorithmic}
\end{algorithm}

\subsection{Tricks}
As we known, training a neural network needs to determine many hyperparameters and also needs to adopt some tricks. However, this network is not hard to train, when selecting the node number, we only need to remember that the number is decreasing layer by layer for encoder. For example, if our input is a 500-dimensional data, then for encoder like view 1\textquotesingle s auto-encoder in Figure 1, the numbers of node is [256, 64, 32], and for decoder is [64, 256], where 32 is the middle-hidden layer's node number and usually we define these layer's node numbers are the same for multiple views even each view's input dimension is different. As for learning rate of optimizer algorithm, for each view's auto-encoder, learning rate usually sets to 0.5 or 0.3, and for supervised network, learning rate usually equals to 0.9. Note that suitable learning rates will let each auto-encoder and supervised network help with each other to accelerate the convergence speed. Additionally, we further emphasize that we will obtain the joint latent representation by mapping each view's latent feature into a same space with a same transformation matrix ${{\bf{w}}_{share}}$, because a same transformation matrix can project different data into a same subspace and that is also the reason why we define all views' latent feature as the same dimension. Lastly, one of the most significant tricks is early stopping, because the training loss of a neural network cannot always decrease, and after a period of time, training loss will not decrease any more and even increase. Therefore, early stopping is necessary that it can stable and accelerate the training process. Such as in a training epoch, we set ${R_1} = {R_2} = 1000$, may in round 400, the loss is lowest but we save the model parameters of round 1000, and we miss the best model parameters and will train more rounds which is a waste of time (because our early stopping rule is that if ${R_1},{R_2} >  = 200$, and the loss no longer decrease for 200 rounds, we will break the loop and save the parameters belong to the round corresponds to the best loss).

\begin{figure}[ht]
\begin{center}
\includegraphics[width=0.8\textwidth]{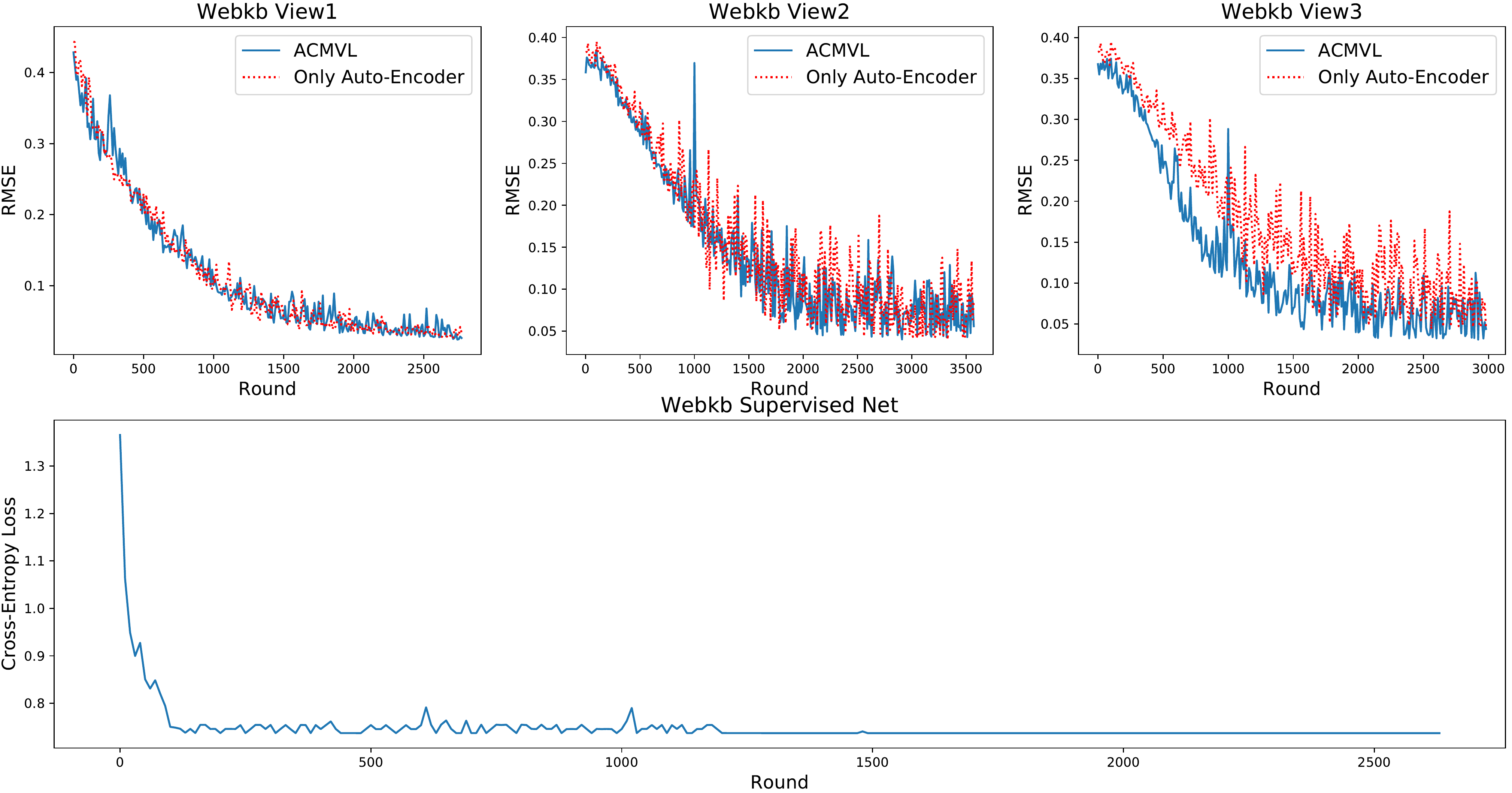}
\caption{Convergence Analysis of WebKb} \label{splncs04}
\end{center}
\end{figure}

\begin{figure}[htbp]
\begin{center}
\includegraphics[width=0.7\textwidth]{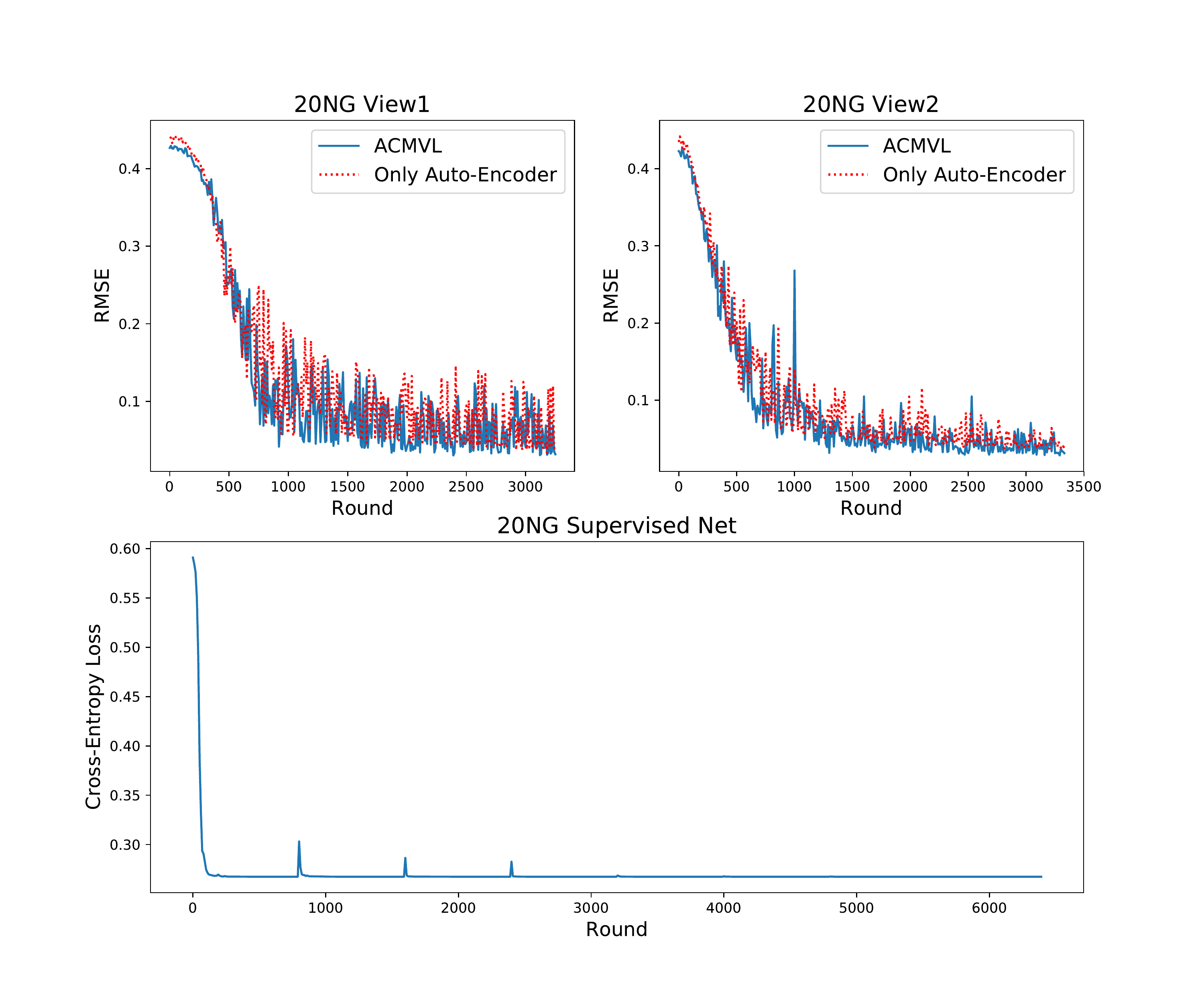}
\caption{Convergence Analysis of 20NG} \label{splncs04}
\end{center}
\end{figure}

\section{Experiment}
\subsection{Data Set Partition}

In this subsection, we give a short description of the data sets and introduce the method to divide the data set into different views.

\textbf{WebKb}: The WebKb data set contains web information from computer science departments of four different universities, obviously, we actually have four data sets, but we compute the average value of four data sets. There are three views in each data set: the words in the main text in each web page of one of the universities is a kind of view; the clickable words in the hyperlinks pointing to other web pages of one of the universities is another view; and the words in the titles of each web page is also a view. On the other hand, there are seven categories in this data set, where we choose four most representative categories in this experiment. In general, we have 3 views in this data set.

\textbf{20NewsGroup}: The data set consists of 20 News group, that is to say, this data set contains 20 categories. 200 documents are randomly selected from each category. As a result, we define 20 tasks corresponding to the classes, and the documents belong to the category related to the task are defined as positive instances, and from other different categories are defined as negative ones. Next, we take the words appearing in all the tasks as a common view, and the words only existing in each task as a special view. In this way, we get 21 views, however, there are only two views in each task, 19 views are missed in each task \cite{DBLP:conf/pkdd/JinZWHS13}. Now this is a multi-task and multi-view data, but we can conduct the experiment on each task and compute the average value of them. In general, we have 2 views in this data set.

\textbf{Leaves}: The leaves data set includes leaves from one hundred plant species that are divided into 32 different genera, and 16 samples of leaves for each plant species are presented. 3 geniuses that have 3 or more plant species are selected to form the data set, and the aim of the problem is to discriminate different species in a genus. And in this data set, three views of features are available, including shape descriptor, fine scale margin and texture histogram, and each view has 64 features. In general, we have 3 views in this data set.
\begin{figure}[ht]
\begin{center}
\includegraphics[width=0.8\textwidth]{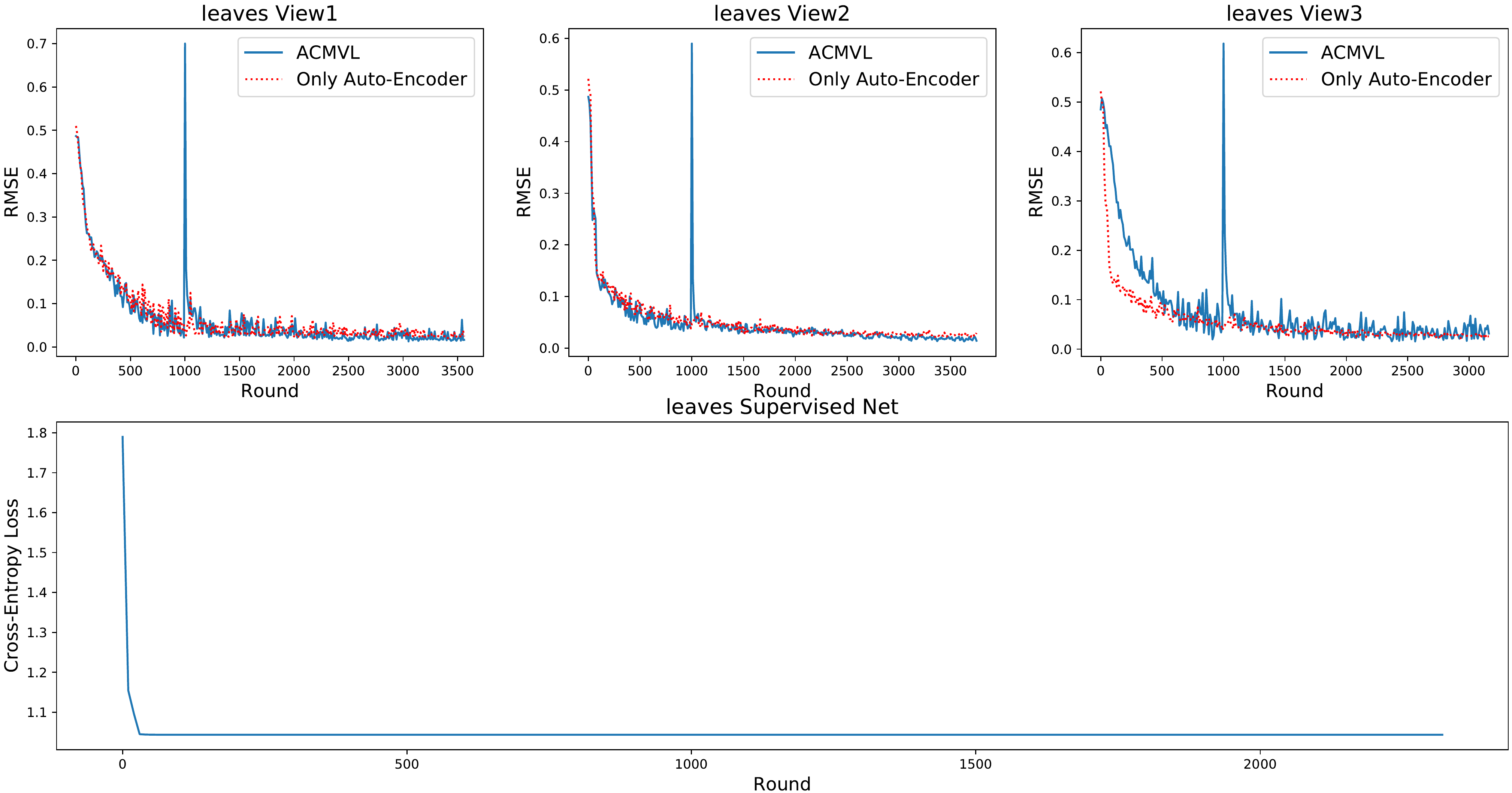}
\caption{Convergence Analysis of Leaves} \label{splncs04}
\end{center}
\end{figure}
\subsection{Convergence Analysis of Training Process}
In Figure 3, Figure 4, and Figure 5, the first row of figures shows each views' reconstruction error of auto-encoder, where, red line shows the original auto-encoder's curve, and the blue one is our proposed method's curve. It's not hard to see that in Figure 2 and Figure 4 let each auto-encoder converges faster than the original auto-encoder, ACML has less fluctuation in training process than original ones. However, we find in Figure 5 original auto-encoder performs better because Leaves is a small data set that only original auto-encoder can fit it soon, but our proposed method will train the model alternately. Recall that after ${R_1}$ rounds in one epoch of training each auto-encoder, we will need to train the supervised network for ${R_2}$ rounds and modify each auto-encoder's parameters of encoder, and then in the next epoch, we should retrain each auto-encoder. In such way, when facing small data, the speed of convergence of ACMVL may be slower than the original auto-encoder's. However, when dealing with bigger data set, ACMVL can accelerate convergence speed of each auto-encoder as shown in Figure 3 and Figure 4.

On the other hand, with the help of each auto-encoder, supervised network converges fast as shown in the second row of Figure 3, Figure 4, and Figure 5. As a result, in ACMVL's framework, each view's auto-encoder and supervised network help with each other according to the co-training rules.

\renewcommand{\arraystretch}{1.5}
\begin{table}[htbp]
\setlength{\abovecaptionskip}{2pt}
  \setlength{\belowcaptionskip}{2pt}
   \centering
   \fontsize{8}{10}\selectfont
   \caption{Classification Task}
   \label{tabl}
\begin{tabular}{c|c|cccccccc}
\hline\hline
\multicolumn{2}{c|}{\multirow{2}{*}{\textit{Classification}}} & \multicolumn{2}{c}{\textit{LR}} & \multicolumn{2}{c}{\textit{LR-AE}} & \multicolumn{2}{c}{\textit{LR-AE-ACMVL}}   & \multicolumn{2}{c}{\textit{LR-ACMVL}}                                        \\ \cline{3-10}
\multicolumn{2}{c|}{}                                         & \textbf{ACC}    & \textbf{F1}    & \textbf{ACC}     & \textbf{F1}     & \textbf{ACC} & \textbf{F1}                 & \textbf{ACC}                              & \textbf{F1}                      \\ \hline\hline
\multirow{3}{*}{\textit{WebKb}}        & \textbf{View 1}      & 0.8230          & 0.7281         & 0.7876           & 0.6217          & 0.8142       & \multicolumn{1}{c|}{0.7330} & \multirow{3}{*}{\textbf{0.9115}} & \multirow{3}{*}{\textbf{0.8703}} \\ \cline{2-8}
                                       & \textbf{View 2}      & 0.8673          & 0.7866         & 0.8142           & 0.6991          & 0.7434       & \multicolumn{1}{c|}{0.6667} &                                           &                                  \\ \cline{2-8}
                                       & \textbf{View 3}      & 0.7080          & 0.6477         & 0.7168           & 0.6031          & 0.7256       & \multicolumn{1}{c|}{0.5839} &                                           &                                  \\ \hline
\multirow{2}{*}{\textit{20NG}}         & \textbf{View 1}      & 0.6267          & 0.6256         & 0.5333           & 0.5326          & 0.5200       & \multicolumn{1}{c|}{0.5169} & \multirow{2}{*}{\textbf{0.9733}}          & \multirow{2}{*}{\textbf{0.9732}} \\ \cline{2-8}
                                       & \textbf{View 2}      & 0.9666          & 0.9667         & 0.8600           & 0.8592          & 0.9200       & \multicolumn{1}{c|}{0.9200} &                                           &                                  \\ \hline
\multirow{3}{*}{\textit{Leaves}}       & \textbf{View 1}      & 1.000           & 1.000          & 1.000            & 1.000           & 1.000        & \multicolumn{1}{c|}{1.000}  & \multirow{3}{*}{\textbf{1.000}}           & \multirow{3}{*}{\textbf{1.000}}  \\ \cline{2-8}
                                       & \textbf{View 2}      & 0.7708          & 0.7481         & 0.7917           & 0.7680          & 0.7708       & \multicolumn{1}{c|}{0.7514} &                                           &                                  \\ \cline{2-8}
                                       & \textbf{View 3}      & 0.9167          & 0.9132         & 0.8750           & 0.8713          & 0.9375       & \multicolumn{1}{c|}{0.9325} &                                           &                                  \\ \hline\hline
\end{tabular}
\end{table}

\subsection{Performance Test on Multiple Learning Task}
In this subsection, we will test the algorithm performance on classification and clustering tasks. We select Logistic Regression (LR) as the baseline method of classification as well as Gaussian Mixture Model (GMM) as the baseline method of clustering.

\textbf{Classification}: First of all, we divide the data of each view into training set and testing set at a ratio of 50\%. And then, we first conduct an experiment only use these data with the classifier of LR, and the result is list in the column LR of Table 1; Second, we only train each view's auto-encoder without the help of supervised network, and then we use the feature computed by auto-encoder which corresponds to the training set to train LR classifier and use the feature corresponding to the testing set to test the result. The result is list in the column LR-AE of Table 1; Thirdly, we will do the same thing as the second experiment but let supervised network helps each auto-encoder's training process, and the result is list in the column LR-AE-ACMVL of Table 1. Note that this experiment use feature computed by each view's auto-encoder to test each view's performance; Lastly, we conduct the third experiment again, but we use the joint latent feature computed by supervised network to test the performance, and therefore, this experiment only has one view. The result is list in the column LR-ACMVL of Table 1.

In classification experiment, we select Accuracy (ACC) and F1 score (F1) as the metrics. We can easily find that when using ACMVL, most of auto-encoders' performance get better and the joint latent feature's performance is much better than each view's, which verify ACMVL is an effective approach to compute joint latent feature of multiple views.

\renewcommand{\arraystretch}{1.5}
\begin{table}[htbp]
\setlength{\abovecaptionskip}{2pt}
  \setlength{\belowcaptionskip}{2pt}
   \centering
   \fontsize{8}{10}\selectfont
   \caption{Clustering Task}
   \label{tabl}
\begin{tabular}{c|c|cccccccc}
\hline\hline
\multicolumn{2}{c|}{\multirow{2}{*}{\textit{Clustering}}} & \multicolumn{2}{c}{\textit{GMM}} & \multicolumn{2}{c}{\textit{GMM-AE}} & \multicolumn{2}{c}{\textit{GMM-AE-ACMVL}}   & \multicolumn{2}{c}{\textit{GMM-ACMVL}}                                        \\ \cline{3-10}
\multicolumn{2}{c|}{}                                         & \textbf{NMI}    & \textbf{JC}    & \textbf{NMI}     & \textbf{JC}     & \textbf{NMI} & \textbf{JC}                 & \textbf{NMI}                              & \textbf{JC}                      \\ \hline\hline
\multirow{3}{*}{\textit{WebKb}}        & \textbf{View 1}      & 0.0935          & 0.1947         & 0.1539           & 0.1460          & 0.3004       & \multicolumn{1}{c|}{0.2088} & \multirow{3}{*}{\textbf{0.5809}} & \multirow{3}{*}{\textbf{0.0310}} \\ \cline{2-8}
                                       & \textbf{View 2}      & 0.4739          & 0.1593         & 0.1221           & 0.0752          & 0.3752       & \multicolumn{1}{c|}{0.0885} &                                           &                                  \\ \cline{2-8}
                                       & \textbf{View 3}      & 0.0843          & 0.1947         & 0.0433           & 0.1858          & 0.0752       & \multicolumn{1}{c|}{0.1549} &                                           &                                  \\ \hline
\multirow{2}{*}{\textit{20NG}}         & \textbf{View 1}      & 0.0195          & 0.5167         & 0.0057           & 0.3567          & 0.0110       & \multicolumn{1}{c|}{\textbf{0.3500}} & \multirow{2}{*}{0.3171}          & \multirow{2}{*}{0.5100} \\ \cline{2-8}
                                       & \textbf{View 2}      & 0.1413          & 0.5133         & 0.1968           & 0.4533          & \textbf{0.4021}       & \multicolumn{1}{c|}{0.4133} &                                           &                                  \\ \hline
\multirow{3}{*}{\textit{Leaves}}       & \textbf{View 1}      & 0.8567          & 0.2917         & 0.8461            & 0.3333           & 0.8642     & \multicolumn{1}{c|}{0.1250}  & \multirow{3}{*}{\textbf{0.9781}}           & \multirow{3}{*}{0.1667}  \\ \cline{2-8}
                                       & \textbf{View 2}      & 0.7215          & 0.3021         & 0.3240           & 0.0938          & 0.7999       & \multicolumn{1}{c|}{\textbf{0.0104}} &                                           &                                  \\ \cline{2-8}
                                       & \textbf{View 3}      & 0.7620          & 0.1042         & 0.8555           & 0.2083          & 0.8024       & \multicolumn{1}{c|}{0.0208} &                                           &                                  \\ \hline\hline
\end{tabular}
\end{table}

\textbf{Clustering}:  We will conduct some semi-clustering experiment, first of all, we divide the data of each view into training set and testing set at a ratio of 50\%. And then, we first conduct an experiment only use testing data with the clustering algorithm GMM, and the result is list in the column GMM of Table 2; Second, we only use training data to train each view's auto-encoder without the help of supervised network, and then we use the model to compute testing data's feature representation and conduct clustering task on these feature representations. The result is list in the column GMM-AE of Table 2; Thirdly, we will do the same thing as the second experiment but let supervised network helps each auto-encoder's training process, and the result is list in the column GMM-AE-ACMVL of Table 2. Note that this experiment use feature computed by each view's auto-encoder to test each view's performance; Lastly, we conduct the third experiment again, but we use the joint latent feature computed by supervised network to test the performance, and therefore, this experiment only has one view. The result is list in the column GMM-ACMVL of Table 2.

In clustering experiment, we select Normalized Mutual Information (NMI) and Jaccard Coefficient (JC, the smaller of JC, the performance of clustering is better) as the metrics. We can easily find that when using ACMVL, most of auto-encoders' performance get better and the joint latent feature's performance is much better than each view's, which verify ACMVL is an effective approach to compute joint latent feature of multiple views.

\section{Conclusion}
In this paper, we propose a novel multi-view learning algorithm called Auto-Encoder based Co-Training Multi-View Representation Learning (ACMVL), which is aimed to subspace learning and model training strategy. We utilize the latent feature learning ability of auto-encoder to grasp the complementarity of multiple views, and at the same time, by using weight sharing we can map each view's latent representation in to a same space and learn the consistency of multiple views. Besides, we adopt co-training strategy to accelerate the training procedure of each view's auto-encoder by co-training and model parameters partially shared. And according to experimental results, we find that our proposed method can learn a suitable joint latent representation which is competent to classification and clustering learning tasks.

Our proposed method in this paper is a deterministic model which cannot measure the uncertainty of latent space. Therefore, in the future, a main target is to find a generative method to obtain the distribution of the joint latent space instead of an instantiation of the space. To our knowledge, variational auto-encoder \cite{DBLP:journals/corr/KingmaW13} may be a good choice to solve this problem. Generally, multi-view subspace learning is a great research direction which is hard but deserved to pay more attention on it, and we will make more attempts and explorations in this field.

%

%
% ---- Bibliography ----
%
% BibTeX users should specify bibliography style 'splncs04'.
% References will then be sorted and formatted in the correct style.
%

%\bibliography{ref}
\bibliographystyle{splncs04}
%\bibliography{mybibliography}

\end{document}